\documentclass[journal]{IEEEtran}

\usepackage{amsmath,amssymb,amsfonts}
\usepackage{cite}
\usepackage{graphicx}
\usepackage{svg}
\usepackage{hyperref}
\usepackage{mathrsfs}
\usepackage{booktabs}
\usepackage{multirow}
\usepackage{stfloats}

\begin{document}
%
\title{FAPP: Fast and Adaptive Perception and Planning for UAVs in Dynamic Cluttered Environments}
%

\author{Minghao Lu$^1$, 
        Xiyu Fan$^1$,
        Han Chen$^2$, 
        and Peng Lu$^{1*}$
\thanks{$^1$Minghao Lu, Xiyu Fan and Peng Lu are with the Adaptive Robotic Controls Lab (ArcLab), Department of Mechanical Engineering, The University of Hong Kong, Hong Kong, SAR, China (e-mail: minghao0@connect.hku.hk; fanxiyu@connect.hku.hk; lupeng@hku.hk)}
\thanks{$^2$Han Chen is with Huawei Technologies Co., Ltd, Wuhan, Hubei, China (e-mail: stark.chen@connect.polyu.hk).}
\thanks{$^*$Corresponding author.}
    \thanks{
        This work was supported by General Research Fund under Grant 17204222, and in part by the Seed Fund for Collaborative Research and General Funding Scheme-HKU-TCL Joint Research Center for Artificial Intelligence.
        }%
}

%



\maketitle

\begin{abstract}
Obstacle avoidance for Unmanned Aerial Vehicles (UAVs) in cluttered environments is significantly challenging. Existing obstacle avoidance for UAVs either focuses on fully static environments or static environments with only a few dynamic objects. In this paper, we take the initiative to consider the obstacle avoidance of UAVs in dynamic cluttered environments in which dynamic objects are the dominant objects. This type of environment poses significant challenges to both perception and planning. Multiple dynamic objects possess various motions, making it extremely difficult to estimate and predict their motions using one motion model. The planning must be highly efficient to avoid cluttered dynamic objects. This paper proposes Fast and Adaptive Perception and Planning (FAPP) for UAVs flying in complex dynamic cluttered environments. A novel and efficient point cloud segmentation strategy is proposed to distinguish static and dynamic objects. To address multiple dynamic objects with different motions, an adaptive estimation method with covariance adaptation is proposed to quickly and accurately predict their motions. Our proposed trajectory optimization algorithm is highly efficient, enabling it to avoid fast objects. Furthermore, an adaptive re-planning method is proposed to address the case when the trajectory optimization cannot find a feasible solution, which is common for dynamic cluttered environments. Extensive validations in both simulation and real-world experiments demonstrate the effectiveness of our proposed system for highly dynamic and cluttered environments.
\end{abstract}

\begin{IEEEkeywords}
Aerial systems, dynamic environment, point cloud, motion planning, obstacle avoidance
\end{IEEEkeywords}

\section*{Supplementary materials}
\textcolor{red}{Video: \url{https://youtu.be/4DXBuKpqQk4}}

%
\IEEEpeerreviewmaketitle

\section{Introduction}
%
%
%
%
\IEEEPARstart{T}{he} development of robotics has always aimed at increasing the intelligence of robots and integrating them into our lives. Unmanned aerial vehicles (UAVs) have gained popularity due to their motion flexibility and cost-effectiveness.
Recent advancements in aerial autonomy technology have enabled UAVs to excel in various intelligent tasks, including navigation, exploration in static environments\cite {racer, egoplanner}, and autonomous cinematography\cite{cinema}. 

However, flying UAVs in complex and dynamic cluttered environments without human command remains a challenging and potentially hazardous endeavor.
For instance, consider a scenario where a UAV is assigned the task of capturing close-up photography on city sidewalks bustling with pedestrians or in narrow indoor spaces filled with crowds. In such situations, the UAV needs to react intelligently to the ever-changing environment while accomplishing its mission.

State-of-the-art obstacle avoidance for UAVs mainly focuses on static environments \cite{richter2016polynomial,racer,egoplanner,Song2023}. Dynamic environments introduce more significant challenges in both perception and path planning. The motion blur caused by dynamic objects, especially fast-moving objects, making them difficult to detect using traditional visual sensors. The path planning algorithms must be efficient such that they can avoid fast-moving objects. As such, event-based detection methods have been developed to address fast-moving objects \cite{falanga2020dynamic}. However, event cameras are expensive and not cost-efficient for low-cost UAVs. Therefore, the challenge of UAVs navigating in dynamic environments remains open. Recently, several methods have been proposed to address dynamic obstacles for UAVs \cite{lin2020robust,he2021fast,fastsmall,dpmpc}. However, all of these studies consider the case where the environment is static with only a few (one or two) dynamic objects. They either use constant velocity or acceleration model \cite{lin2020robust,he2021fast,CHEN2022104124,fastsmall,dpmpc} to estimate and predict the object's motion, or let the object walk with a constant velocity \cite{wang2021autonomous}. They did not investigate whether the estimation is fast or precise and its effects on the success of dynamic obstacle avoidance.

In this paper, we consider a more complex environment: a dynamic cluttered environment where dynamic objects are the dominant objects in the environment. This type of environment will aggregate the difficulties in perception and planning. For perception, detecting multiple objects and tracking them is much more challenging than detecting one object. For dynamic obstacle avoidance, it is necessary to estimate and predict the velocity of the objects. This also becomes significantly more challenging as different objects possess different motion models. It is difficult to use one model to estimate the motions of different objects. For planning, it is also more challenging as multiple objects increase the probability of collisions. Furthermore, it may be common that the planning may fail to find a feasible solution in dynamic cluttered environments. All of these challenges make the obstacle avoidance in a dynamic cluttered environment significantly challenging.


This paper proposes FAPP (Fast and Adaptive Perception and Planning) for UAVs in dynamic cluttered environments. We first propose a novel simple but highly efficient point cloud segmentation strategy that can efficiently distinguish static and dynamic objects. A unique data association method is developed to assign detected dynamic clusters to existing clusters. To address multiple dynamic objects, a novel adaptive estimation is proposed to quickly estimate and predict their velocities, which facilitates the avoidance of multiple objects. In terms of path planning, we further improve our previous planning \cite{fastsmall} without even using the front-end path searching and safe corridor generation. Furthermore, an adaptive re-planning strategy is proposed to address the situation where a feasible path cannot be found, which is common in dynamic cluttered environments. Finally, we perform various simulations and real experimental flight tests to validate the performance of our proposed FAPP, both indoors and outdoors. In summary, the main innovations of this paper are as follows:
\begin{itemize}
\item[1)] A novel and efficient point cloud segmentation strategy is proposed, which can efficiently distinguish static and dynamic objects.

\item[2)] A novel covariance adaptation method is proposed to address multiple dynamic objects with different motions. This method can overcome the limitation of the constant velocity or acceleration assumption made by existing studies. It can quickly estimate and predict the positions and velocities of the objects, which facilitates the avoidance of multiple dynamic objects.


\item[3)] An adaptive re-planning method is proposed to address the situation when no feasible path can be found by the trajectory optimization. This is important for a dynamic cluttered environment in which no feasible solution is common.


\item[4)] To the best of our knowledge, this is one of the first few works that consider the obstacle avoidance of UAVs in highly cluttered and dynamic environments. We validate the performance of our FAPP in various simulation and experimental tests. The whole perception and planning process can be completed within a few milliseconds, which is highly efficient.

\end{itemize}

\section{Related Works}
Obstacle avoidance and planning in complex dynamic environments continue to present significant challenges. Existing works have identified two main components of the problem: dynamic object perception and mapping, and vehicle motion planning. However, developing a complete system that can effectively handle real-time perception and obstacle avoidance of any dynamic object remains a formidable task.

\subsection{Dynamic environment perception}
For the optimal response of a UAV's motion in a random dynamic environment, precise recognition of the environment is crucial. This includes accurate mapping of the scene and effective segmentation and motion estimation of dynamic objects within the environment. Some recent works use the event-based method\cite{falanga2020dynamic,he2021fast} to detect moving obstacles for obstacle avoidance. However, event cameras are only sensitive to dynamic targets and it is difficult to build a static local map of the environment in real-time. In \cite{fastsmall,activesense}, the image-based algorithm is used for moving object detection and tracking. However, the current limitations exist where only specific objects can be detected, and the detection is not coupled with mapping capabilities. Point cloud-based methods can better handle the problem but are still difficult to meet the requirements of real-time obstacle avoidance. In \cite{DynamicFilter,erasor,remove}, the researchers proposed methods to remove dynamic points from the global map, but cannot instantly and accurately detect dynamic objects. Dynamic object segmentation by clustering and motion estimation is also extensively studied over years\cite{CHEN2022104124,wang2021autonomous,lin2020robust,leveraging}. However, in these works, the description of dynamic objects, static objects, or occluded objects is not comprehensive enough, which limits their practicality to more complex scenarios. Some learning-based methods can achieve moving object segmentation more accurately \cite{automatic2022,receding2022}, but they are difficult to run in real-time on mobile terminals without GPUs.

\subsection{Dynamic obstacle avoidance}
The obstacle avoidance problem for robots, especially for UAVs, has been widely studied and explored in recent years. Some reactive methods, including velocities obstacles(VO)\cite{fiorini1998motion,CHEN2022104124}, artificial potential field(APF)\cite{hybrid2017,falanga2020dynamic,he2021fast} have been used for UAV dynamic obstacle avoidance. The approaches only take consideration of current moving obstacles and only compute one-step actions, which can not meet the requirements in environments with dense hybrid (static and dynamic) obstacles. Model predictive control (MPC) is also widely utilized for dynamic obstacle avoidance \cite{lin2020robust, dpmpc}, as it can achieve optimality in the local time domain. However, these approaches lack the advantage of UAVs' navigation in clustered environments due to their limited description of the complex structures within the environments. Several complete solutions based on optimization in the community have demonstrated fast, robust flights in a clustered static environment, and the motion primitives can be solved in real time \cite{fastplanner, richter2016polynomial, egoplanner}. Inspired by the above works, \cite{wang2021autonomous, fastsmall} adds the penalty of the dynamic obstacles for the polynomial trajectory optimization. However, \cite{wang2021autonomous} did not consider the uncertainty of the moving object prediction and the control effort of the trajectory, while the polygon generation in \cite{fastsmall} is complex and very time-consuming in a cluster environment. Moreover, none of the above work considers how the UAV should react autonomously when it is unable to plan to the target point.
Due to these problems, these techniques are not flexible and may lead to a decrease in performance in a complex dynamic environment.

Overall, all these studies on dynamic obstacle avoidance only consider the environment where only one or two dynamic objects are present and none of them has considered the dynamic cluttered environments in which dynamic objects are the dominant objects.

Furthermore, the motion of the dynamic obstacles considered in the existing studies is rather simple. They either let the object move with a constant velocity \cite{wang2021autonomous} or use a constant velocity or acceleration model \cite{lin2020robust,CHEN2022104124,fastsmall} to estimate and predict their motions. This can hardly be true for dynamic cluttered environments in which many objects possess various motions that does not satisfy the constant velocity or acceleration assumption. Another important issue missing in these existing studies is that they did not show how fast and accurate the estimation and prediction of the object's motion is. Actually, the speed and accuracy of the estimation plays an important role in the success of dynamic obstacle avoidance.

\section{Fast and Adaptive Perception of Dynamic Cluttered Environments}
In this section, we will present the first part of FAPP: fast and adaptive perception. Our fast and adaptive perception is a point-cloud-based method to obtain the real-time static local map while segmenting and estimating the motion of the dynamic objects in the environment. The pipeline of the system is shown in Fig.1.

\begin{figure}[h]
\centerline{\includegraphics[scale=0.57]{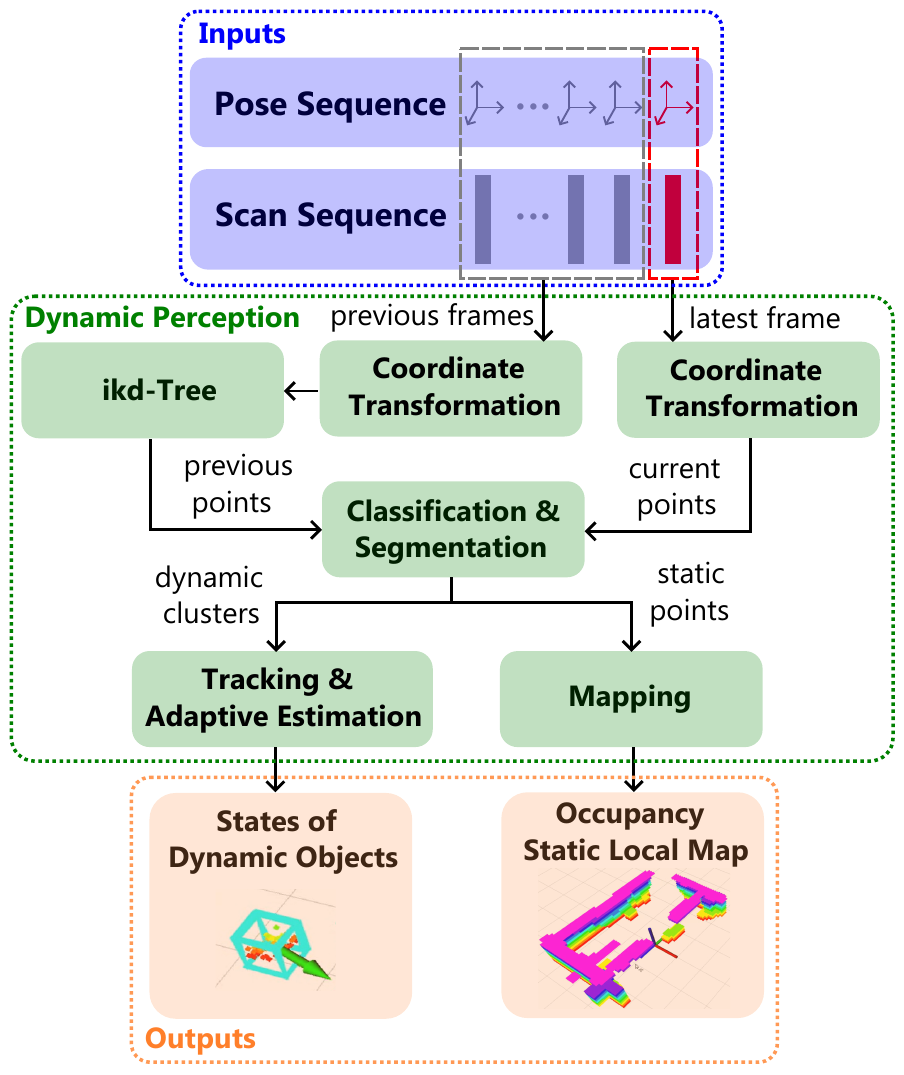}}
\caption{The pipeline of proposed fast and adaptive perception algorithm. The inputs of the algorithm are the sequences of lidar scans and robot poses, and the outputs are static local maps and the states of dynamic objects.}
\label{fig}
\end{figure}

\subsection{Preliminary}
Assuming at the $k$th timestamp $t_{k}$, we can get a 3D point cloud $_{B}P_{k}$ from the scan of the sensor in the local frame. We first make an $SE(3)$ transformation $_{B}^{W}T_{k}$ from the local sensor frame $B$ to the global frame $W$, and get $_{W}P_{k}$. $_{B}^{W}T_{k}$ can be obtained by using the LIO algorithm such as \cite{fastlio}. Then, we create an incremental kd tree (I-KD Tree) \cite{cai2021ikd} to maintain the latest previous $F$ frames point set $\xi$ by adding and deleting points dynamically, where $\xi \subseteq \{\underset{j \in [k-F,k-1]}{\bigcup} {_{W}P_{j}}\}$. Using the I-KD Tree, we can keep the point set $\xi$ a proper size and make the points in $\xi$ uniformly distributed in 3D space, which gives a good representation of the spatial distribution information of obstacles over the previous few frames scan.

\subsection{Fast Dynamic Object Segmentation}
For dynamic obstacle avoidance, it is important to identify dynamic obstacles. While there exists studies that distinguish static objects and dynamic objects using learning-based approaches, they are computationally intensive and difficult to use for small low-cost UAVs.
State-of-the-art UAV dynamic obstacle avoidance filters the point cloud and then clusters it into objects using DBSCAN \cite{dbscan}. By designing a Kalman filter for each cluster, they can detect dynamic objects by estimating their speed \cite{lin2020robust,wang2021autonomous,CHEN2022104124}. This method is effective for sparse environments. However, this is significantly computationally intensive for cluttered environments with many objects. In this paper, we propose a fast dynamic object detection and tracking algorithm. We first propose a novel efficient algorithm that can segment static and dynamic objects without using neural networks. Then, we only design a tracker for dynamic objects. By doing this, we can efficiently detect and track dynamic objects in cluttered environments.

In this subsection, we propose a novel and efficient algorithm to identify the point cloud as dynamic or static. First, we use DBSCAN to cluster the point cloud of the current frame $_{W}P_{k}$, resulting in a set of $m$ clusters $\mathcal C_{k} = \{ C^{1}, C^{2}, ..., C^{m}\}$. For each cluster, we can categorize it into one of three cases:

$\bullet$ Case 1: Continuously moving object.

$\bullet$ Case 2: Static object.

$\bullet$ Case 3: Unknown object.

For case 3, the unknown object encompasses static objects that were previously occluded by moving objects as well as newly appeared objects within the field of view (FOV). Here, we measure the global nearest distance $d^{n}$ from each point $p_{n}$ in a cluster $C^{n}$ with $N$ points in total to the previous point set $\xi$. Then, we propose to formulate the following two functions to describe the classification of a cluster:
\begin{align}
\mathcal T_{1} &= \frac{1}{N} \sum\limits_{n=1}^{N} d^{n}, \label{e:T1}\\
 \quad \mathcal T_{2} &= \frac{1}{N} \sum\limits_{n=1}^{N} \frac{(d^{n} - \mathcal T_{1})^{2}}{\mathcal T_{1}^{2}} \label{e:T2}
\end{align}
where $\mathcal T_{1}$ represents the average minimum distance from the points in the cluster to the previous point set, and $\mathcal T_{2}$ represents the normalized average variance of $d^{n}$. For a static cluster that has been observed previously, the average distance $T_{1}$ from its points to the previous cloud should be within a small measurement error. For a continuously moving cluster, the average distance $\mathcal T_{1}$ should be large, and the nearest point in $\xi$ of each point in the cluster is the projection of itself that once observed. So the distribution of $d^{n}$ will be relatively uniform, while an occluded object or a newly observed object will have an uneven distribution. Thus, the cluster can be classified by two constant thresholds $h_{1}$ and $h_{2}$ as follows:

$\bullet$ Case 1 (Moving objects): $\mathcal T_{1} > h_{1} \quad \& \quad \mathcal T_{2} < h_{2}$ 

$\bullet$ Case 2 (Static objects): $\mathcal T_{1} < h_{1}$

$\bullet$ Case 3 (Unknown objects): $\mathcal T_{1} > h_{1} \quad \& \quad \mathcal T_{2} > h_{2}$ 

\begin{figure}[h]
\centerline{\includegraphics[scale=0.46]{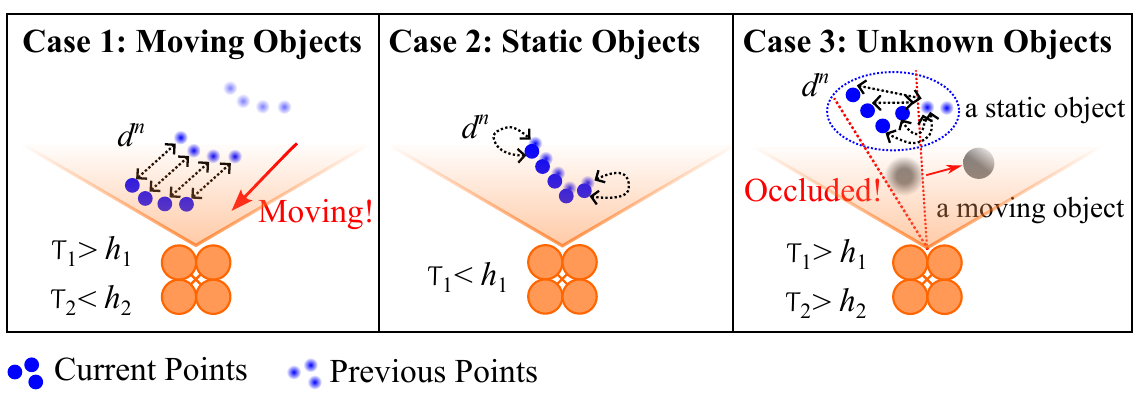}}
\caption{The illustration of our dynamic object segmentation method. For a continuously moving object, its point cloud is at a greater distance from the previous point cloud set, while the distance distribution is relatively uniform.}
\label{fig}
\end{figure}

The description of the process is presented in Fig.2. With the algorithm above, we achieve the point cloud clustering and segmentation of the dynamic objects. The set of $D$ dynamic clusters at time $k$ is $^{d}\mathcal C_{k}$, and the geometric centre of the $d$th cluster can result in $\mathbf o_{k}^{d} = [o_{x}^{d}, o_{y}^{d}, o_{z}^{d}]_{k}^{\mathrm T}$. The overall dynamic points group in the current frame can be written as $_{W}P_{dyn,k}$, which satisfies:
\begin{equation}
_{W}P_{dyn,k} = \underset{d \in [1,D]}{\bigcup} ^{d}\mathcal C_{k}, \enspace _{W}P_{dyn,k} \subseteq \enspace _{W}P_{k},
\end{equation}

Our proposed segmentation algorithm with the proposed two functions in Eqs.~\eqref{e:T1} and \eqref{e:T2} is highly efficient. It can used in highly cluttered environments.
The time efficiency and segmentation results will be demonstrated and compared with state-of-the-art methods in Section VI.

\subsection{Dynamic Object Tracking}
For dynamic obstacle avoidance, it is necessary to track and estimate the motion states of the moving objects after segmentation. The state estimation and prediction will be used in fast and adaptive planning in Section V.

\emph{1) Motion estimation}: Here we describe our motion estimation method. The inter-frame displacements
of each object in the world frame are approximated as a constant velocity model in this work, and we adopt the Kalman Filter to estimate the states of the moving objects. We define the state vector of an object as: $X^{i} = [x_{i}, y_{i}, z_{i}, \dot x_{i}, \dot y_{i}, \dot z_{i}]^{\mathrm T}$, so the system model of each object can be formulated as:
\begin{equation}
X^{i}_{k+1} = A_{k} X^{i}_{k} + \nu^{i},
\end{equation}
\begin{equation}
Z^{i}_{k} = H_{k} X^{i}_{k} + \omega^{i},
\end{equation}
\begin{equation}
A_{k} = \left[\begin{array}{ccc} I_{3\times3} & \Delta t\cdot I_{3\times3} \\ 0_{3\times3} & I_{3\times3} \end{array}\right], H_{k} = \left[\begin{array}{ccc} I_{3\times3} & 0_{3\times3} \\ 0_{3\times3} & I_{3\times3} \end{array}\right],
\end{equation}
where $A$ is the state transition matrix, $H$ is the measurement matrix, $\nu \sim \mathcal N(0,Q)$ is the normally distributed process noise with a covariance matrix $Q$ and $\omega \sim \mathcal N(0,R)$ is the normally distributed measurement noise with a covariance matrix $R$. $\Delta t$ is the time interval between two adjacent frames of scans. If a cluster is newly observed, a new Kalman Filter tracker will be created, while if a cluster can be associated with the $i$th tracker, we use the geometric center and its frame differencing of the cluster as the measurement $Z^{i} = [o_{x}^{i}, o_{y}^{i}, o_{z}^{i}, \Delta o_{x}^{i}, \Delta o_{y}^{i}, \Delta o_{z}^{i}]^{\mathrm T}$. Then, the Kalman Filter can predict and update in real time. The output is the estimated state $\hat{X^{i}}$. 



\emph{2) Data association}: In assigning detected dynamic objects to existing trackers, we propose a data association method for point cloud.
First, we create a $D \times I$ assignment matrix $\mathcal H$ for the $D$ detected objects and $I$ trackers. The matrix is expressed based on the Mahalanobis distance $\Omega_{d,i}$ between each detection $\mathbf o$ and predicted position $\mathbf{\hat{o}}$ from existing trackers:
\begin{equation}
\Omega_{d,i} = \sqrt{(\mathbf o^{d} - \mathbf{\hat{o}}^{i})^{T} \sigma_{i}^{-1} (\mathbf o^{d} - \mathbf{\hat{o}}^{i})},
\end{equation}
\begin{equation}
\mathcal H_{d,i} = 1-  \frac{2}{\pi} \text{arctan}(\Omega_{d,i}),
\end{equation}
where $d=(0,1,...,D)$ and $i=(0,1,...,I)$. $\sigma_{6\times 6}$ is the covariance matrix between the detection and prediction, which can be obtained by the update process of Kalman Filter. 
$\mathcal H_{d,i} \in (0,1)$ indicates the relevance while a larger value denotes a better match. The assignment is solved optimally using the Hungarian algorithm similar to \cite{deepsort}. Additionally, the match of a pair of detection and tracker will also be rejected if the cost value is less than a constant threshold $th_{min}$. 

\subsection{Static Local Map Output}
In the above, we obtain all clusters of the dynamic points of the environment. Next, we eliminate the dynamic points $_{W}P_{dyn,k}$ from the current frame point cloud $_{W}P_{k}$ and obtain $_{W}P_{sta,k}$, which represents the static points in the environment. The process can be represented as:
\begin{equation}
_{W}P_{sta,k} = \enspace _{W}P_{k} - \enspace _{W}P_{dyn,k},
\end{equation}
The static point cloud $_{W}P_{sta,k}$, which includes the complete information of the free space, is used to build an occupancy grid map that will be used for the obstacle avoidance.

\section{Adaptive estimation and prediction for dynamic cluttered environments}
For dynamic obstacle avoidance, it is important to estimate and predict the motion of the dynamic objects. In dynamic cluttered environments, there exist multiple dynamic objects with different motions. Existing studies on UAV dynamic obstacle avoidance assume that the velocity or acceleration of the object is constant \cite{lin2020robust,CHEN2022104124,fastsmall} or let the people walk with a constant velocity \cite{wang2021autonomous}, and then estimate and predict the position and velocity. The constant velocity or acceleration model works when the velocity or acceleration of the dynamic object does not change rapidly. However, multiple dynamic objects have all sorts of motion (one may suddenly change its direction while maintaining the speed) and it is difficult to use one model to estimate and predict their motion. One solution in dynamic object tracking is to use multiple models with each model representing one motion assumption (constant velocity or constant acceleration). However, this will significantly increase the computational burden as we need an estimator for every dynamic object.

In this paper, we propose to estimate and predict the motion of dynamic objects using a covariance adaptation method without the use of multiple models. We still use the constant velocity model as mentioned in subsection III.C. This model will yield a wrong estimation when the velocity of the objects changes rapidly as the process model is not consistent with the real motion model. Consequently, the innovation of the Kalman filter will significantly deviate from zero-mean. Therefore, we propose to use the innovation sequence to update the covariance of the process noise covariance matrix $Q_k$ to obtain a better estimation and prediction.

For the Kalman Filter of each dynamic object, we can compute its innovation sequence at time step $k$ as follows:
\begin{align}
\gamma_k = Z_k - H_k \hat{X}_{k|k-1}
\end{align}
where $\hat{X}_{k|k-1}$ denotes the prediction (a prior estimation) of $X_k$. By using the innovation sequence, we can compute its actual covariance as follows:
\begin{align}
C_{\gamma,k} = \frac{1}{W} \sum\limits_{l=k-W+1}^{k}  \gamma_l \gamma_l^{\mathrm T}
\end{align}
where $W$ is the size of the sliding window. Meanwhile, we can also compute the theoretical innovation covariance using the Kalman Filter. By making use of the covariance matching property, we can obtain the following:
\begin{align}
H_{k} \hat{Q}_{k} H_{k}^{\mathrm T}= C_{\gamma,k} - H_{k}A_{k-1} P_{k-1|k-1} A_{k-1}^{\mathrm T} H_{k}^{\mathrm T}  - R_{k} 
\end{align}
Then, we could obtain the estimate $\hat{Q}_{k}$ as follows:
\begin{align}
\hat{Q}_{k} = C_{\gamma,k} - A_{k-1} P_{k-1|k-1} A_{k-1}^{\mathrm T}   - R_{k}
\end{align}
where $P_{k-1|k-1}$ denotes the estimation error covariance matrix at the previous time step $k-1$. Finally, to guarantee the positiveness of the covariance matrix \cite{Lu2016a}, we use the following matrix to update $Q_k$ to prevent numerical errors:
\begin{equation}
{Q}_{k} = \mathrm{diag}(\mathrm{max}\{0,\widetilde {Q}_{k}(1,1)\},...,\mathrm{max}\{0,\widetilde {Q}_{k}(6,6)\}).
\end{equation}

By updating the process noise covariance matrix $Q_k$ using the adaptation method, we can obtain a fast and satisfactory estimation of the object's motion even when its motion changes significantly, which will be shown in Section VI. To quickly and accurately estimate and predict the motion of the dynamic objects is vital for dynamic obstacle avoidance. Slow and wrong estimation could easily lead to collisions with the dynamic object. However, existing studies \cite{lin2020robust,wang2021autonomous,CHEN2022104124,fastsmall} ignored this important issue. We will demonstrate in Section VI that a fast and precise prediction of the dynamic objects plays the key role for obstacle avoidance in complex dynamic environments.

%

\section{Fast and Adaptive planning}
In this section, we will present our fast and adaptive planning which contains a trajectory optimization and an adaptive re-planning strategy. Our trajectory optimization does not require a front-end path search and takes into account the uncertainty of dynamic objects' state estimation. Additionally, an adaptive re-planning is developed to address the case when a feasible solution is not found. The output is a dynamically feasible and safe trajectory.

\subsection{Trajectory Definition}
For a differentially flat system such as the UAV, we first define its motion as a piece-wise 3-dimension and 5-degree polynomials $p(t)$ with $M$ pieces, and the $l$th piece can be expressed as:
\begin{equation}
\label{7}
p_{l}(t) = \mathbf{c}_{l}^{\mathrm T} \beta(t), \quad t \in [0,T_{l}],
\end{equation}
where $\mathbf{c}_{l} \in \mathbb{R}^{6 \times 3}$ is the coefficient matrix of the piece and $\beta(t) = [1, t, ..., t^{5}]^{\mathrm T}$ is natural basis vector. $T_{l}$ is the duration of the piece.

Then, we adopt $\mathbf{MINCO}$ (minimum control) \cite{minco} class to achieve the spatial-temporal decoupled optimization. The trajectory $p(t)$ can only be parameterized by the time duration of each piece $\mathbf T = [T_{1}, ..., T_{M}]^{\mathrm T}$ and the intermediate waypoints $\mathbf q = [q_{1}, ..., q_{M-1}]^{\mathrm T}$ with a convenient space-time deformation $\mathcal M$:
\begin{equation}
\mathbf c = \mathcal M(\mathbf q, \mathbf T),
\end{equation}
where $\mathbf c = [\mathbf c_{1}^{\mathrm T}, ..., \mathbf c_{M}^{\mathrm T}]$. 

\subsection{Problem Formulation}
For $\mathbf{MINCO}$, a general optimization problem can be formulated as:
\begin{equation}
\min\limits_{\mathbf{q},\mathbf{T}} \mathcal J(\mathbf{q},\mathbf{T}) + \sum {\lambda_{\star}} \mathcal I_{\star}(\mathbf{c}(\mathbf{q},\mathbf{T}),\mathbf{T}),
\end{equation}
\begin{equation}
\mathcal J = \int _{0}^{T_{t}} ||p^{(3)}(t)||^{2} \mathrm{d}t + \rho T_{t}, \quad T_{t} = \sum\limits_{l=1}^{M} T_{l},
\label{e:J}
\end{equation}
\begin{equation}
\mathcal I_{\star} = \sum\limits_{l=1}^{M} \frac{T_{l}}{\kappa_{l}} \sum\limits_{\tau = 0}^{\kappa_{l}} \mathcal{G}_{\star}(\mathbf{c}_{l},T_{l},\frac{\tau}{\kappa_{l}}),
\end{equation}
where $\mathcal J$ is the time-regularized control effort, which minimizes the jerk and the total duration $T_{t}$ of the trajectory, and $\rho$ is the tunable weight value. $\mathcal I$ is the time integral penalty with weight $\lambda$, and $\kappa$ means the sample numbers on a piece of trajectory, $\frac{\tau}{\kappa}$ indicates the specific timestamp. 

Then, we expect different types of cost function $\mathcal G_{\star}$ to indicate the requirements of dynamic obstacle avoidance. The detailed penalty functions are designed as follows:

\emph{1) Dynamical feasibility}: The motion of the robot has to satisfy the dynamical feasibility. Here, we limit the range of velocity and acceleration by the cost function $\mathcal{G}_{f}$ expressed as follows:
\begin{equation}
\mathcal{G}_{v} = \max \{ (||p_{l}^{(1)}(t_{\tau})||^{2} - v_{max}^{2}), 0 \}^{3},
\end{equation}
\begin{equation}
\mathcal{G}_{a} = \max \{ (||p_{l}^{(2)}(t_{\tau})||^{2} - a_{max}^{2}), 0 \}^{3},
\end{equation}
\begin{equation}
\mathcal{G}_{f} = \mathcal{G}_{v} + \mathcal{G}_{a},
\end{equation}
where $v_{max}$ and $a_{max}$ are the maximum allowed magnitudes of velocity and acceleration. $t_{\tau} = T_{l} \frac{\tau}{\kappa_{l}}$ is a specific sample timestamp on the $l$th piece of the trajectory.

\emph{2) Static obstacle avoidance}: In Section III, we have obtained the static occupancy grid map, and the static obstacles can be modeled by the method in \cite{egoplanner}. Then, for any point of $p(t)$, its distance to the closest obstacle can be calculated as a function $d_{s}(p(t))$. Then, the static obstacle collision cost $\mathcal{G}_{s}$ can be formulated as:
\begin{equation}
\mathcal{G}_{s} = \max \{ (\mathcal{D}_{s} - d_{s}(p_{l}(t_{\tau}))) , 0 \}^{3},
\end{equation}
where $\mathcal{D}_{s} = r_{0} + \epsilon$ is the safety threshold between the robot and the obstacle surface, meaning a distance that a small constant $\epsilon$ greater than the radius of the robot $r_{0}$.

\emph{3) Dynamic obstacle avoidance}: Based on the tracking method of the moving objects proposed in \uppercase\expandafter{\romannumeral3}.C, the predicted trajectory $p_{b}(t)$ of a moving object with the estimated state position $\mathbf{\hat{o}}_{k}^{i}$ and velocity $\mathbf{\hat{\dot{o}}}_{k}^{i}$ at time $t_{k}$ can be represented as:
\begin{equation}
p_{b}^{i}(t) = \mathbf{\hat{o}}_{k}^{i} + \mathbf{\hat{\dot{o}}}_{k}^{i} (t - t_{k}),
\end{equation}

By doing this, we can evaluate the safety of a trajectory at any time stamp. Therefore, for $I$ tracked moving objects, the dynamic obstacle collision cost $\mathcal G_{d}$ can be designed as:
\begin{equation}\mathcal{G}_{d} = \sum\limits_{i=1}^{I} \max \{ ({\mathcal{D}_{d}^{i}}^{2} - ||p_{l}(t_{\tau}) - p_{b}^{i}(t_{\tau})||^{2}) , 0 \}^{3},
\end{equation}
where $\mathcal D_{d}^{i} = r_{0} + r^{i} + e^{i}$ is the safety clearance between the robot with radius $r_{0}$ and the $i$th object with radius $r^{i}$. The component $e^{i}$ represents the uncertainty of the estimated position at $t_{\tau}$, which can be defined by:
\begin{equation}
e^{i} = \sqrt{\sigma_{\tau}^{i}(1,1)^{2} + \sigma_{\tau}^{i}(2,2)^{2} + \sigma_{\tau}^{i}(3,3)^{2}},
\end{equation}
where $\sigma_{\tau}^{i} = A_{\tau}\sigma_{k}^{i}A_{\tau}^{\mathrm T}$, means the transition of the covariance matrix from timestamp $t_{k}$ to $t_{\tau}$ of the object state. The difference between the formulation of $A_{\tau}$ and $A_{k}$ is $\Delta t$ in $A_{k}$ and $(t_{\tau}-t_{k})$ in $A_{\tau}$. The optimization for dynamic obstacles is described in Fig. 3.
\begin{figure}[h]
\centerline{\includegraphics[scale=0.55]{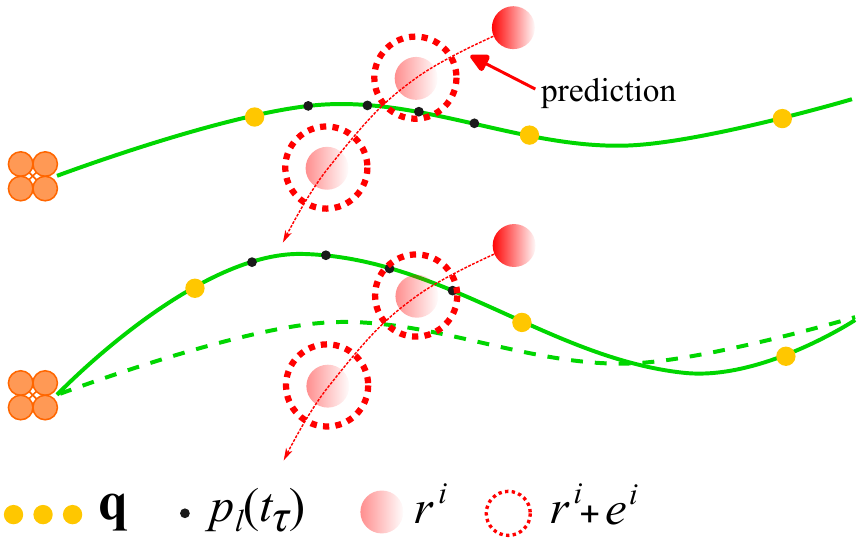}}
\caption{The optimization of the trajectory to avoid a dynamic object. When the trajectory is not safe, the penalty will be added at each sample point (black dots), and the penalty incorporates the uncertainty of the prediction of dynamic objects (red dotted circle). After the optimization, the trajectory will be collision-free with dynamic objects. The optimization changes the position of intermediate waypoints $\mathbf{q}$ (yellow dots), as well as the time allocation of the trajectory pieces.}
\label{fig}
\end{figure}

Considering the increasing uncertainty of $p_{b}^{i}$ with the propagation of prediction time, $\mathcal{G}_{d}$ is only valid while the prediction time $(t-t_{k})$ is within a given range.

Above all, we have obtained the overall penalty in Eq.~\eqref{e:J}, and write it as $J(\mathbf{q},\mathbf{T})$ for simplicity. To solve the optimization problem, we need the gradient of $J$ w.r.t $\mathbf q$ and the gradient of $J$ w.r.t $\mathbf T$, which can be derived by the Gradient Propagation Law:
\begin{equation}
\frac{\partial J(\mathbf{q}, \mathbf{T})}{\partial \mathbf{q}}=\frac{\partial J}{\partial \mathbf{c}} \frac{\partial \mathbf{c}}{\partial \mathbf{q}},\ 
\frac{\partial J(\mathbf{q}, \mathbf{T})}{\partial \mathbf{T}}=\frac{\partial J}{\partial \mathbf{T}}+\frac{\partial J}{\partial \mathbf{c}} \frac{\partial \mathbf{c}}{\partial \mathbf{T}},
\end{equation}
where $\partial \mathbf{c} / \partial \mathbf{q}$ and $\partial \mathbf{c} / \partial \mathbf{T}$ can be derived from (8). Hitherto, this problem can be solved efficiently by unconstrained optimization algorithms such as L-BFGS.

\subsection{Adaptive Re-planning Strategy}
Our trajectory optimization can usually find a feasible solution when the environment is not complex. However, in the presence of dynamic cluttered environments, especially when the environment is small with many dynamic objects, trajectory optimization can usually fail. One common solution is to let the drone stay put to prevent wrong decisions. This decision, however, is dangerous in dynamic cluttered environments as dynamic objects can easily collide with the drone. To address this issue, we propose an adaptive re-planning strategy to allow our drone to continue to carry out tasks without collisions even when a feasible solution is not found.

Generally, our autonomous UAV has two mission states: navigating and hovering.
Here, we present our adaptive planning strategy in the two states as follows: 

\emph{1) While navigating}: When the UAV is navigating to a target position $\mathbf{p}_{f}\in \mathbb{R}^{3}$, it is necessary to check the collision with static and dynamic obstacles within a certain period using the UAV trajectory $p(t)$, moving objects trajectory $p_{b}(t)$ and the occupancy grid map. If a collision risk is detected, the UAV will replan the trajectory based on the current state. However, if the dynamic obstacles are too dense, the feasible collision-free trajectory to the target point $\mathbf{p}_{f}$ may not exist. In this case, we need to calculate a temporary target position $\mathbf{p}_{f}^{*}$. First, for a moving object with a velocity $\mathbf{v}_{s}$ and a relative position vector $\mathbf{r}_{s}$ to UAV, we generate a repulsion force vector $\mathbf{n}_{s}$ where satisfies:
\begin{equation}
\mathbf{n}_{s} = \frac{\mathbf{v}_{s} \cdot \mathbf{r}_{s}}{||\mathbf{r}_{s}||^2} \mathbf{r}_{s},
\end{equation}
where $\mathbf{n}_s$ indicates the component vector of $\mathbf{v}_s$ in the direction of $\mathbf{r}_{s}$. Then, we can calculate the total repulsion force vector $\mathbf{n}_{total} = \sum\limits_{s=1}^{S} \mathbf{n}_{s}$ for the $S$ objects. Finally, $\mathbf{p}_{1}$ can be determined by:
\begin{equation}
\mathbf{p}_{f}^{*} = p(t_{r}) + h \frac{\mathbf{n}_{total}}{||\mathbf{n}_{total}||},
\label{e:pstar}
\end{equation}
where $h$ is the horizon of the intermediate target distance, and $t_{r}$ is the time stamp while conducting the replanning. The description of this process is shown in Fig. 4.

\begin{figure}[h]
\centerline{\includegraphics[scale=0.75]{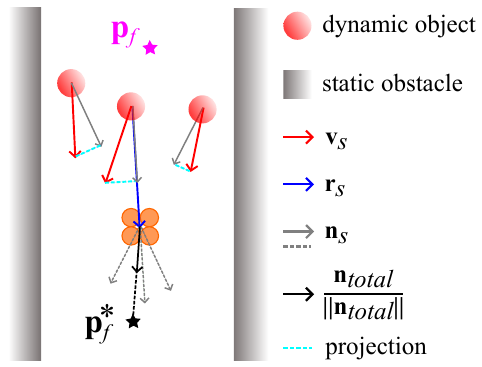}}
\caption{The principle of the autonomous strategy. The method will plan a temporary target when the feasible collision-free trajectory to the target point $\mathbf{p}_{f}$ does not exist. The repulsion force vector $\mathbf{n}_{s}$ pushes the UAV to free space. A larger projection of $\mathbf{v}_s$ in the direction of $\mathbf{r}_s$ results in a more significant repulsion effect.}
\label{fig}
\end{figure}

\emph{2) While hovering}: If the UAV is hovering at $\mathbf{p}_{0} \in \mathbb{R}^{3}$ and waiting for targets, while $S$ objects are moving toward the UAV that has a risk of collision in a short time $\delta$, we expect the UAV to fly away from the objects by determining an intermediate target point $\mathbf{p}_{f}^{*}$ to get a trajectory and replan to the original position after avoiding successfully. After the objects move away in $\delta$, the robot will replan the trajectory back to the origin. The calculation of $\mathbf{p}_{f}^{*}$ is the same with Eq.~\eqref{e:pstar}.

\section{Experiments and Evaluations}

In this section, we test our system in both simulated and real-world dynamic cluttered environments. Comprehensive quantitative analyses are conducted to evaluate the advantages of our proposed system. 

\subsection{Implementation Details}

For the experiments, we designed our experimental hardware platform UAV290, which is a 290mm wheelbase frame with the protection of the rotors, carrying an Intel NUC12WSHi7 running Ubuntu 20.04 as the onboard computer, and a Livox MID-360 Lidar with the FOV of 360°(horizontal)$\times$59°(vertical) and detection range of 40m within 10\% reflectivity is equipped for onboard sensing, while publishing point cloud at 50Hz. The controller of the UAV is a PixRacer-Pro running the PX4 flight stack. The overall system weighs 1.96 kg, with dimensions being 450$\times$450$\times$150 mm. The overview of our hardware platform is shown in Fig. 5. The simulation and evaluation of our algorithms are also conducted on Intel NUC12WSHi7. 

\begin{figure}[h]
\centerline{\includegraphics[scale=0.55]{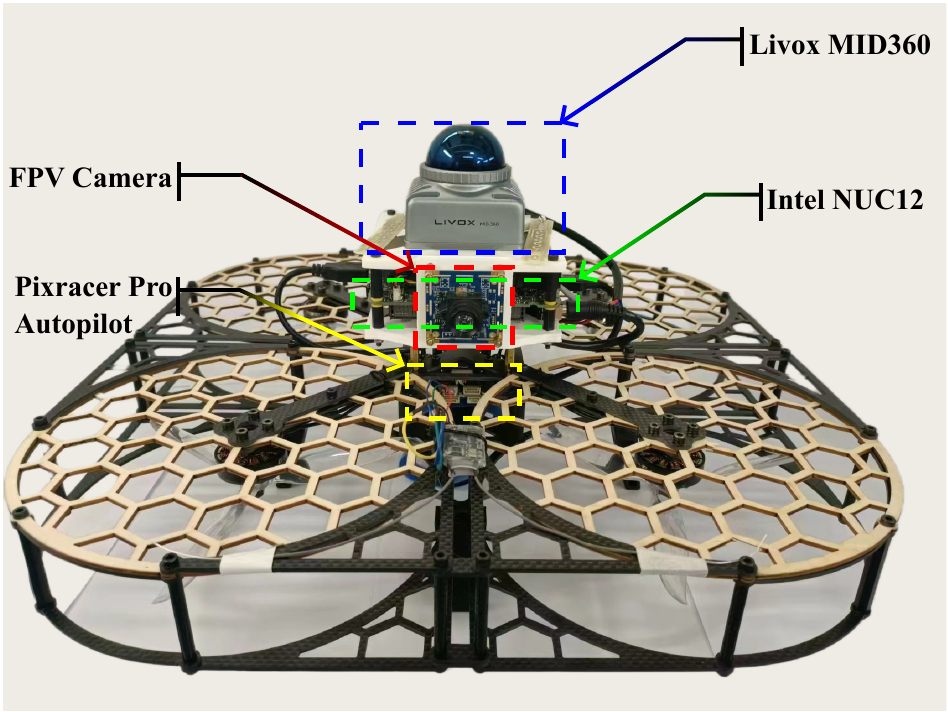}}
\caption{Our UAV290 Hardware Platform.}
\label{fig}
\end{figure}

\subsection{Evaluation of Dynamic Perception}

Firstly, we quantitatively verify the accuracy and stability of our segmentation and state estimation for dynamic objects. 
Our work is compared with SOTA works and the results can be found in Table 1. The mean absolute error of our estimated position $e_{pos}$ achieves 0.11 m and the mean absolute error of our estimated velocity $e_{vel}$ achieves 0.19 m/s. Multiple Object Tracking Accuracy (MOTA) $(\%)$ is defined in \cite{mota}, while a higher value indicates better performance in detecting the dynamic objects and keeping the trackers. For our MOTA, it is composed of a false negative rate $f_{n} = 4.3\%$ (indicating non-detected dynamic objects), a false positives rate $f_{p} = 5.2\%$ (static objects misclassified as dynamic), and a mismatch rate $f_{m} = 6.4\%$. The average time cost $t_{per}$ of our algorithm achieves 12.77 ms, which is practical for real-time running on mobile terminals. 
The evaluation illustrates that our method improves tracking accuracy and robustness with less time cost in the clustered environment, and satisfies the requirements of real-time dynamic obstacle avoidance.
The data collection for this evaluation is conducted in a motion capture room, with a running MID-360 Lidar and 3 people moving at about 1 m/s.

\begin{table}[h]
\centering
\caption{Dynamic Perception Comparison}
\begin{tabular}{ccccc}
   \toprule
   Method & $e_{pos}$($m$) & $e_{vel}$($m/s$) & $\mathbf{MOTA}$(\%) & $t_{per}$($ms$) \\
   \hline
   Ours                      & \textbf{0.17}  & \textbf{0.29}       & \textbf{84.10}      & \textbf{12.77}  \\
   \cite{CHEN2022104124}     & 0.24  & 0.35       & 82.90      & 29.52  \\
   \cite{wang2021autonomous} & 0.33  & 0.37       & 76.40      & 40.33  \\
   \cite{lin2020robust}      & 0.34  & 0.41       & 70.20      & 40.52  \\
   \bottomrule 
\end{tabular}
\label{t:perception comp}
\end{table}

\begin{figure}[t]
\centerline{\includegraphics[scale=0.6]{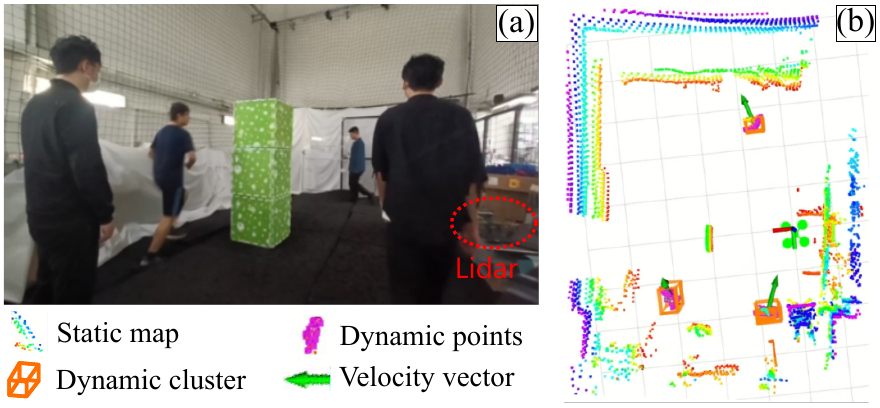}}
\caption{Dynamic perception in a small dynamic cluttered room. (a) is the third-person view image of the scenario. (b) is the data visualization in rviz.}
\label{f:perception1}
\end{figure}

To further demonstrate the effectiveness and efficiency of our proposed segmentation algorithm, we design two more scenarios. In the first scenario, we place the lidar in a room with four people. There are some obstacles surrounding the room and in the middle of the room to test the perception performance when an object is occluded by obstacles. All four persons have significantly different motions: one person standing still, one person walking back and forth, one person walking in a circular motion, and one person running. Please refer to the video for the real-time perception. Even though four people have different motions, their motions are all correctly detected and estimated. One snapshot of the video can be found in Fig. \ref{f:perception1}. It is seen in the figure that all dynamic objects have been segmented from the static local map. There is no false detection of dynamic objects.

\begin{figure}[t]
\centerline{\includegraphics[scale=0.6]{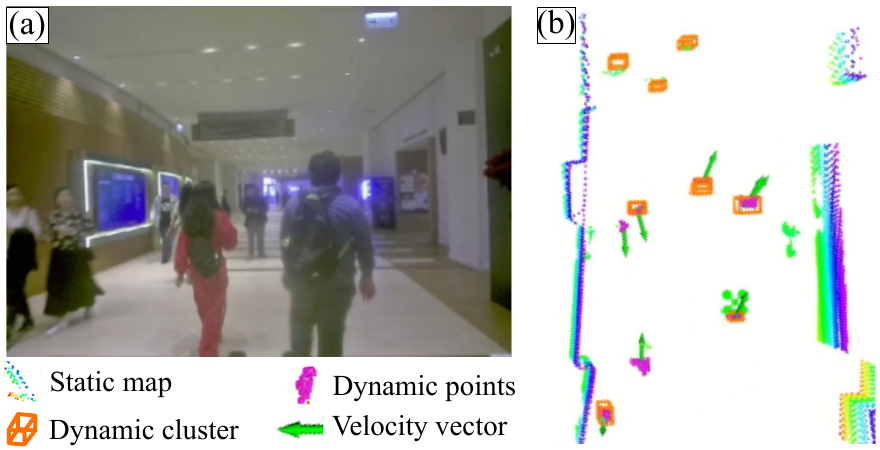}}
\caption{Dynamic perception in a large-scale dynamic cluttered public zone. (a) is the first-person view image of the scenario. (b) is the data visualization in rviz.}
\label{f:perception2}
\end{figure}

Finally, we hand-hold our drone with lidar and walk in a large-scale cluttered public zone. The zone is surrounded by walls, different static objects, and pedestrians. This is to test the perception performance of our algorithm in dynamic cluttered environments. As the zone and the number of objects are large, it is difficult to run existing methods in realtime as they need to design a Kalman filter for every object. Our method first segments the static and dynamic objects and then only design a Kalman filter for dynamic objects, which significantly reduces the computational load. As can be seen in our video, as the lidar moves in the large zone, all dynamic objects are segmented from static objects and their motions are also estimated in real time using our fast and adaptive perception method. One snapshot can be found in Fig. \ref{f:perception2}. It is seen that there are 10 dynamic objects near the lidar. No static walls or objects have been detected as dynamic objects, which shows the superior segmentation performance of our algorithm.

%

\subsection{Evaluation of the Adaptive Perception with Covariance Adaptation}
To show the advantages of our proposed adaptive perception in dynamic cluttered environments, we perform an ablation study. 
In the simulation, we compare the velocity estimation performance with and without covariance adaptation. We simulate three dynamic objects with different motions \emph{1)}: an emerging object with a constant velocity of 5 m/s. \emph{2)}: an object that changes the direction of motion rapidly, with an acceleration at $3m/s^2$ from 0 s to 1.0 s and 1.2 s to 2.0 s, $-30m/s^2$ from 1.0 s to 1.2 s. For this object, the acceleration also changes significantly. Therefore, the constant acceleration model also cannot reflect its real motion. \emph{3)}: an object moving with a sinusoidal velocity, with a period of 1 s and amplitude of 6.28 m/s. 

\begin{figure*}[t]
\centerline{\includegraphics[scale=0.84]{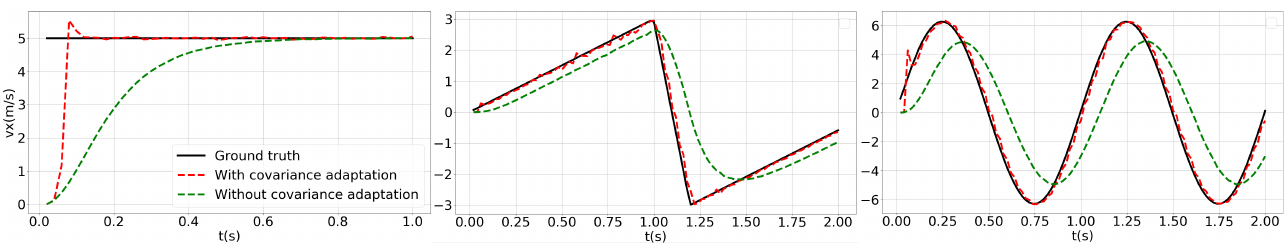}}
\caption{The comparison of the velocity estimation with covariance adaptation and without covariance adaptation.}
\label{f:estimation adapt}
\end{figure*}

The result is shown in Fig. \ref{f:estimation adapt}. As can be seen in Fig. \ref{f:estimation adapt}, the filter with covariance adaptation can quickly estimate the changes in the velocities while the one without covariance adaptation takes a long time to estimate the velocities. The covariance adaptation method can obtain satisfactory results no matter if the velocity or the acceleration changes significantly. The mean absolute errors of the estimations are shown in Table II. We notice that the mean absolute error $e_{vel}$ and the time cost of converging to within 10\% error $t_{con}$ significantly reduce with our covariance adaptation, which demonstrates a higher accuracy and faster response. Especially, when the object is in non-uniform motion, the velocity cannot be predicted well without the covariance adaptation.

\begin{table}[h]
\centering
\caption{Ablation Study of the Covariance Adaptation}
\begin{tabular}{cccc}
   \toprule
   Scenario & Method & $e_{vel}$($m/s$) & $t_{con} (s)$ \\
   \hline
   \multirow{2}{*}{Condition 1} 
        & Ours & \textbf{0.29} & \textbf{0.10}  \\ 
        & w/o adaptation & 0.99 & 0.40 \\ 
    \hline
    \multirow{2}{*}{Condition 2} 
        & Ours & \textbf{0.08} & \textbf{0.08}  \\ 
        & w/o adaptation & 0.56 & / \\ 
    \hline
    \multirow{3}{*}{Condition 3} 
        & Ours & \textbf{0.36} & \textbf{0.14}  \\ 
        & w/o adaptation & 2.39 & / \\ 
   \bottomrule 
\end{tabular}
\end{table}

\begin{figure}[t]
\centerline{\includegraphics[scale=0.5]{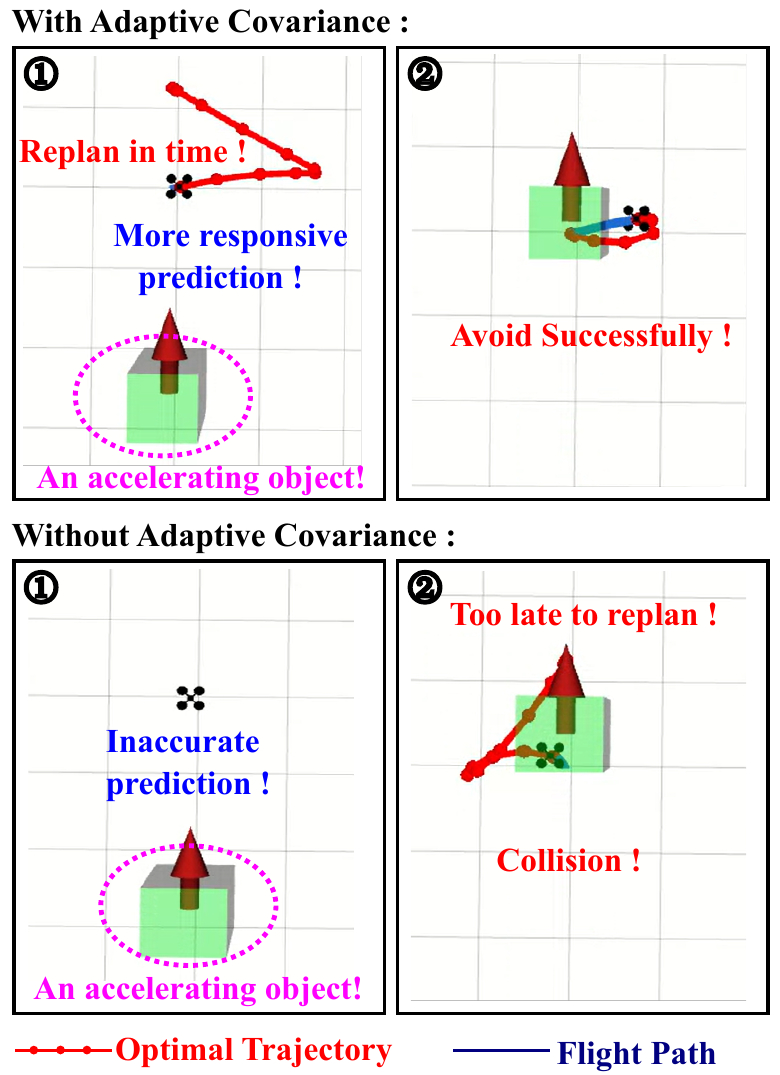}}
\caption{With adaptive covariance updating, the state estimation Kalman filtering will approach the ground truth value faster, which has a significant improvement on obstacle avoidance.}
\label{f:covariance avoidance}
\end{figure}

This improvement in state estimation of dynamic objects will greatly improve the dynamic obstacle avoidance. Without such estimation, obstacle avoidance can easily fail in highly dynamic cluttered environments. We also design multiple simulation tests to show its effectiveness for dynamic obstacle avoidance. We let an object move towards a hovering UAV with a random rapid acceleration. The initial velocity of the object is $1m/s$, and the acceleration of the object ranges from $1m/s^2$ to $5m/s^2$. In 50 tests, the UAV achieves a success rate of 94\% with covariance adaptation, however, when the covariance adaptation is eliminated, the success rate is only 48\%. An example of the comparison is shown in Fig. \ref{f:covariance avoidance}. Due to this reason, for all the following simulation tests or experiments, we all use the perception with the covariance adaptation unless specified.

\subsection{Evaluation of Obstacle Avoidance in Simulation}

\begin{figure}[t]
\centerline{\includegraphics[scale=0.55]{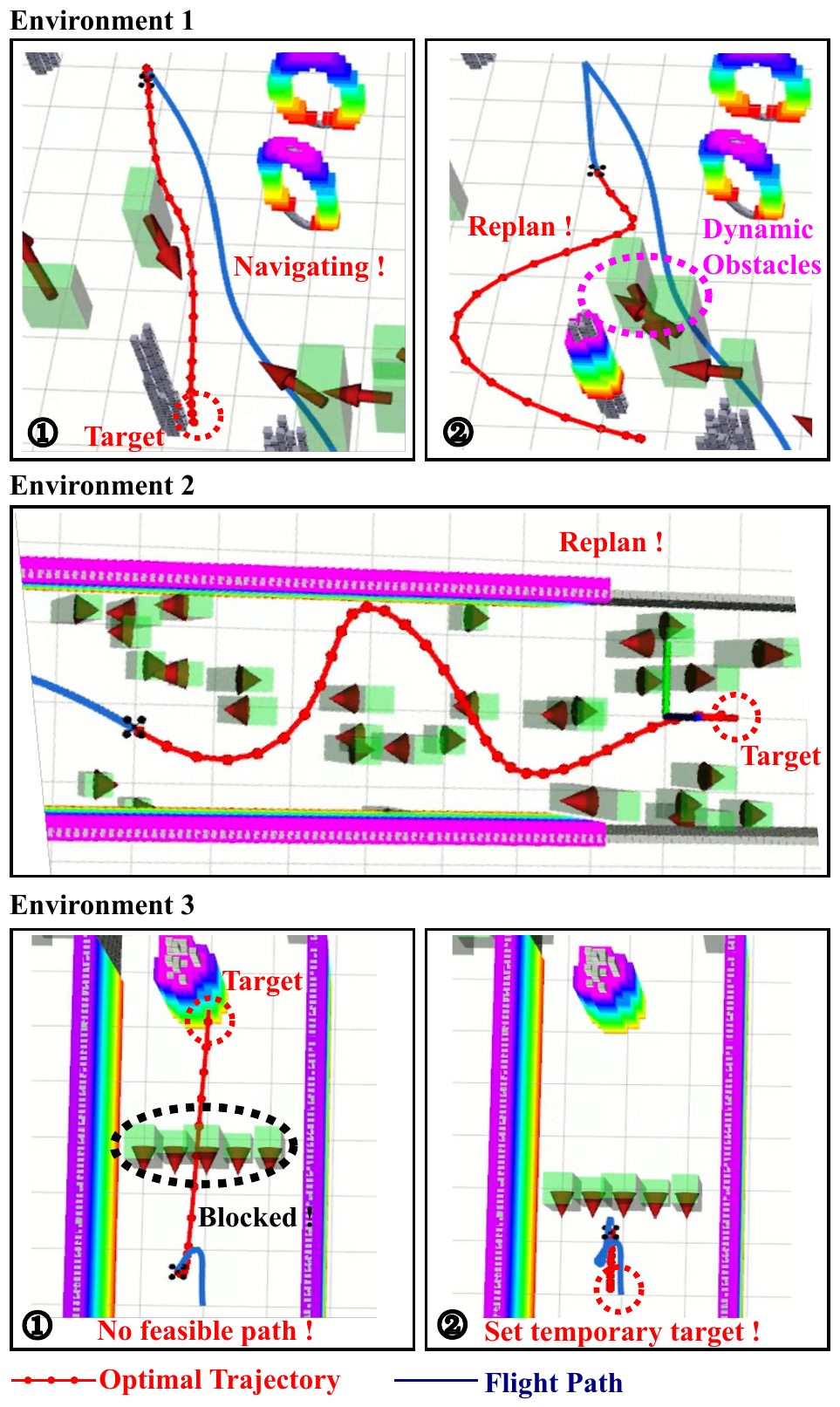}}
\caption{Dynamic planning of UAV in 3 types of simulation environments. }
\label{f:simulation}
\end{figure}

In this section, we present the simulation results of the obstacle avoidance. We demonstrate UAV obstacle avoidance performance in three different environments: \emph{1)}: A field with complex random static obstacles and dynamic objects. The field is 50 m*50 m, with 100 static boxes and 100 static circles randomly placed in it. We generate 100 moving obstacles with random velocities varying from $0.5m/s$ to $3m/s$ and random radius varying from 0.2 m to 1.0 m. The UAV should navigate to any target point safely in this environment. \emph{2)}: A narrow corridor with dynamic objects moving in two opposite directions. The corridor is 40 m in length and 3 m in width, with 50 moving obstacles at velocities varying from 0.5 m/s to 3 m/s. The UAV is required to fly through the corridor without collision. \emph{3)}: A narrow corridor blocked by dynamic objects moving together. 5 moving objects in a row moving at the same velocity $(0.6\ m/s)$ along the corridor. In this case, no feasible collision-free trajectory can be solved to reach the target point, and the UAV is expected to choose a temporary target point for planning. The map and the states of the objects are published at 50Hz. To simulate the time cost of perception, we add the preassigned time delay $t_{delay}$ after the planner receives the objects’ state. $t_{delay}$ is set as 12.77 ms, which is the average time cost of our mapping algorithm shown in Table \ref{t:perception comp}. The visualization in Rviz of the results is shown in Fig.~\ref{f:simulation}. Please refer to our video for the whole dynamic obstacle avoidance process. The performance of the simulation flight tests demonstrates the ability of our method to tackle various complex dynamic cluttered environments.

Afterward, to verify the advantage of our obstacle avoidance method, we compare it with the method in \cite{fastsmall}, \cite{CHEN2022104124}, and \cite{lin2020robust}. We performed 50 tests with each method in the three environments mentioned above and summarized the success rate $\eta (\%)$, average energy cost $E_A (m/s^3)$, and average time cost $t_{plan} (ms)$ of algorithms. When a collision occurs, the test will be considered a failure. Energy cost is the jerk integral of the complete trajectory. The result can be found in Table \ref{t:planning comp}. 

From Table \ref{t:planning comp} we conclude that our method shows superiority in time cost and success rate of obstacle avoidance. Especially in environment 3, only our method can successfully tackle the problem in this situation. The trajectory energy cost of our method is slightly higher than the method in \cite{fastsmall}. This is due to the trajectory in \cite{fastsmall} being initialized by its front-end kinodynamic path searching. However, the polygon generation in the front end will be very time-consuming when there is an excessive density of obstacles. That is why our method is significantly faster than all other methods. Moreover, the constraint of the complex polygons will also limit the feasibility of the trajectory, which might cause a failure of optimization.

\begin{table}[h]
\centering
\caption{Dynamic Planning Benchmark Comparison}
\begin{tabular}{ccccc}
   \toprule
   Scenario & Method & $E_{A}$($m/s^{3}$) & $t_{plan}$($ms$) &  $\eta (\%)$ \\
   \hline
   \multirow{3}{*}{Environment 1}   
                              & Ours                      & 1.44   & 2.48   & 94 \\
                              & \cite{fastsmall}          & 1.23   & 9.59   & 78  \\     
                              & \cite{CHEN2022104124}     & 2.85   & 5.79   & 70  \\     
                              & \cite{lin2020robust}      & 2.14   & 7.21   & 74  \\
   \hline
   \multirow{3}{*}{Environment 2}   
                              & Ours                      & 1.48   & 3.12   & 88  \\
                              & \cite{fastsmall}          & 1.31   & 13.10  & 64  \\    
                              & \cite{CHEN2022104124}     & 2.52   & 6.59   & 58  \\     
                              & \cite{lin2020robust}      & 2.33   & 7.68   & 52  \\
   \hline 
   \multirow{3}{*}{Environment 3}   
                              & Ours                      & 2.35   & 2.80   & 90  \\
                              & \cite{fastsmall}          & /   & /   & /  \\  
                              & \cite{CHEN2022104124}     & /   & /   & /  \\     
                              & \cite{lin2020robust}      & /   & /   & /  \\
   \bottomrule 
\end{tabular}
\label{t:planning comp}
\end{table}

\subsection{Real-world Flight Experiments}
Based on the simulation and quantitative studies, we finally validate our whole system in real-world experiments. Among all tests, the UAV uses the lidar inertial odometry algorithm Fast-Lio2 at 50Hz\cite{fastlio} for localization. The run time of each module in real flight is summarized in Table \ref{t:runtime real}. Overall, the entire system only takes about 20 ms in each iteration.

\begin{table}[h]
\centering
\caption{The Run Time of Each Module of Our Proposed System}
\begin{tabular}{ccc}
   \toprule
   Modules & Time($ms$) & Portion(\%) \\
   \hline
   Onboard Localization              & 4.58    & 22.79  \\
   Fast and Adaptive Perception               & 12.77   & 63.53  \\
   Fast and Adaptive Planning        & 2.75    & 13.68  \\
   $\mathbf{Total}$       & 20.10   & 100    \\
   \bottomrule 
\end{tabular}
\label{t:runtime real}
\end{table}

\begin{figure}[t]
\centerline{\includegraphics[scale=0.43]{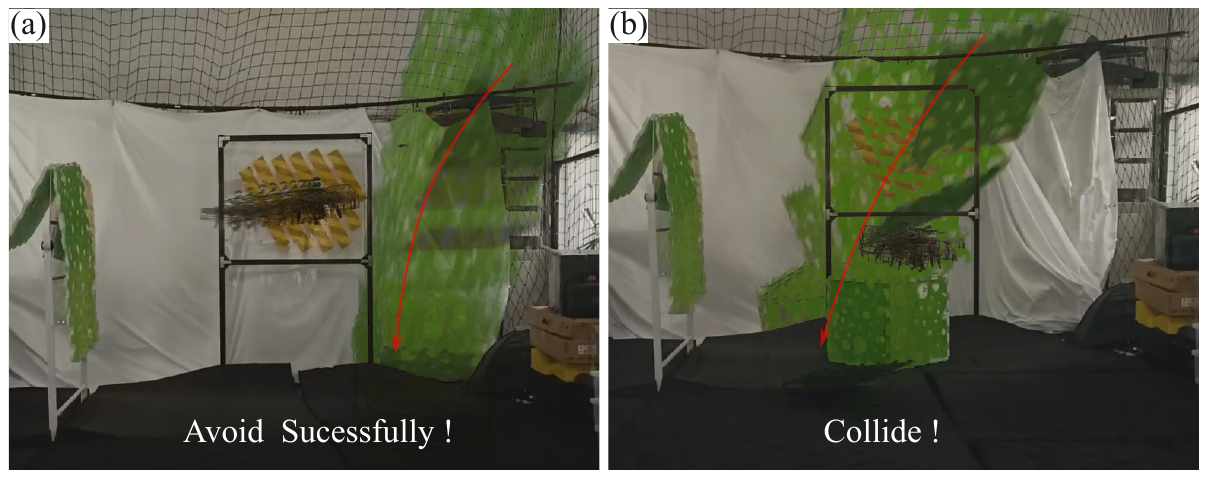}}
\caption{The composed images of dodging a flying box. The left image is a success case running with covariance adaptation, the left is a fail case without covariance adaptation.}
\label{f:covariance real}
\end{figure}

We first performed the ablation experiments to validate the advantage of our covariance adaptation approach. In 10 tests for each of the two groups, people at a distance of about 5 m threw a box toward the hovering UAV. Dynamic obstacle avoidance with covariance adaptation succeeded 9 times, while the one without covariance adaptation succeeded only 2 times.
We present a demo of comparison in Fig.~\ref{f:covariance real}. For this reason, all the following experiments are performed using the covariance adaptation.

\begin{figure*}[h]
\centerline{\includegraphics[scale=0.42]{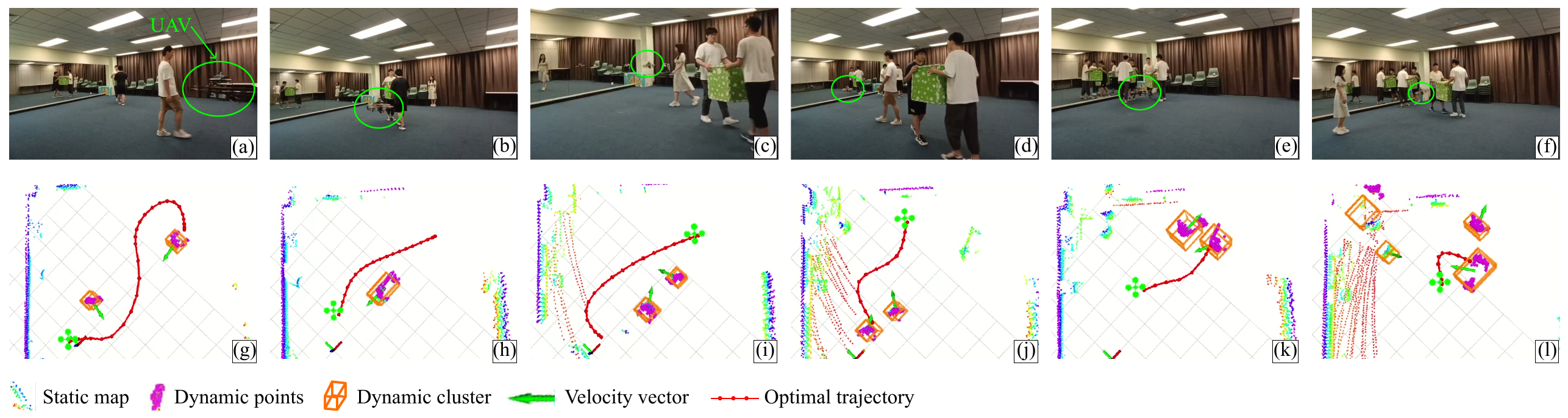}}
\caption{The results of the indoor flight test. (a) - (f) is the third-person view snapshots of the indoor experiment with the corresponding data visualization in (g) - (l). 6 workers were carrying boxes, and the UAV could patrol the small room autonomously.}
\label{f:indoor real}
\end{figure*}

Afterward, we designed two types of scenarios for testing our system for dynamic cluttered environments. In the first scenario, the UAV is required to patrol a room from one corner to another while there are many ``workers'' moving boxes in the patrol route and pedestrians passing through the patrol route. As the UAV size is large, the room filled with so many people is a very cluttered environment with dynamic obstacles. Please refer to the video for the whole avoidance process. Moving ``workers'' and pedestrians are segmented from the static map and their motions are estimated using the adaptive estimation. The system outputs an optimized trajectory based on the predicted states of each dynamic object. The UAV, ``workers'' and pedestrians can work together without collisions. Some snapshots are given in Fig.~\ref{f:indoor real}. 

\begin{figure*}[h]
\centerline{\includegraphics[scale=0.42]{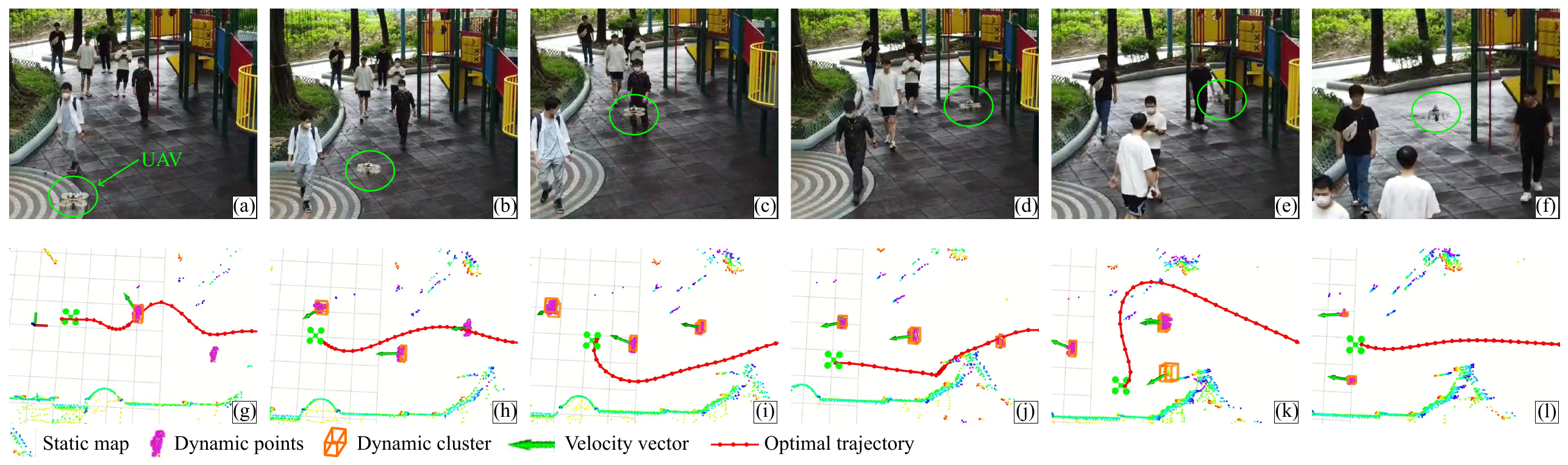}}
\caption{The results of the outdoor flight test. (a) - (f) presents the bird’s-eye view snapshots of the outdoor experiment. (g) - (l) is the screenshots of data visualization in Rviz corresponding with (a) - (f). The UAV avoided the 6 walking persons while navigating on the park sidewalk. }
\label{f:outdoor real}
\end{figure*}

In the second scenario, the UAV is flying on a park sidewalk full of pedestrians and surrounded by common facilities and trees. As can be seen from the video, the UAV can successfully pass all of the pedestrians without collisions with people and the surrounding facilities. Some snapshots are shown in Fig.~\ref{f:outdoor real}. All the experiments are repeated many times. Repeated experiments show similar patterns and are not shown. All these experimental tests verified our proposed system can handle highly dynamic and cluttered environments.

%

\section{Conclusion}
In this paper, we considered the obstacle avoidance of UAVs in a complex dynamic and cluttered environment. We proposed FAPP to tackle the challenges in perception and planning brought by dynamic cluttered objects. The fast perception system can efficiently segment static and dynamic objects. To address the limitation of a constant velocity or acceleration model, we proposed an adaptive estimation that can quickly and accurately predict the motion of multiple dynamic objects. The proposed adaptive estimation also greatly facilitated the dynamic obstacle avoidance. Furthermore, our fast and adaptive planning can even address the case when trajectory optimization cannot find a feasible solution, which is common in dynamic cluttered environments.

Our proposed system performs satisfactorily in dynamic cluttered environments. However, if the onboard localization fails while the UAV is performing aggressive maneuvers, it may experience failure occasionally. Future work would explore how to avoid failures even when the onboard localization fails.


\section*{Acknowledgement}
The authors gratefully thank Huibin Zhao, Qiyuan Qiao, Mingyang Li, Yuting Tao, Xiao Cao, Jingjan Lin, Yi Luo, Peiyu Chen, Weipeng Guan, and Fuling Lin for the experiment support.


%





\ifCLASSOPTIONcaptionsoff
  \newpage
\fi



%



\bibliographystyle{ieeetr}
\bibliography{ref}

%

\begin{IEEEbiography}[{\includegraphics[width=1in,height=1.25in,clip,keepaspectratio]{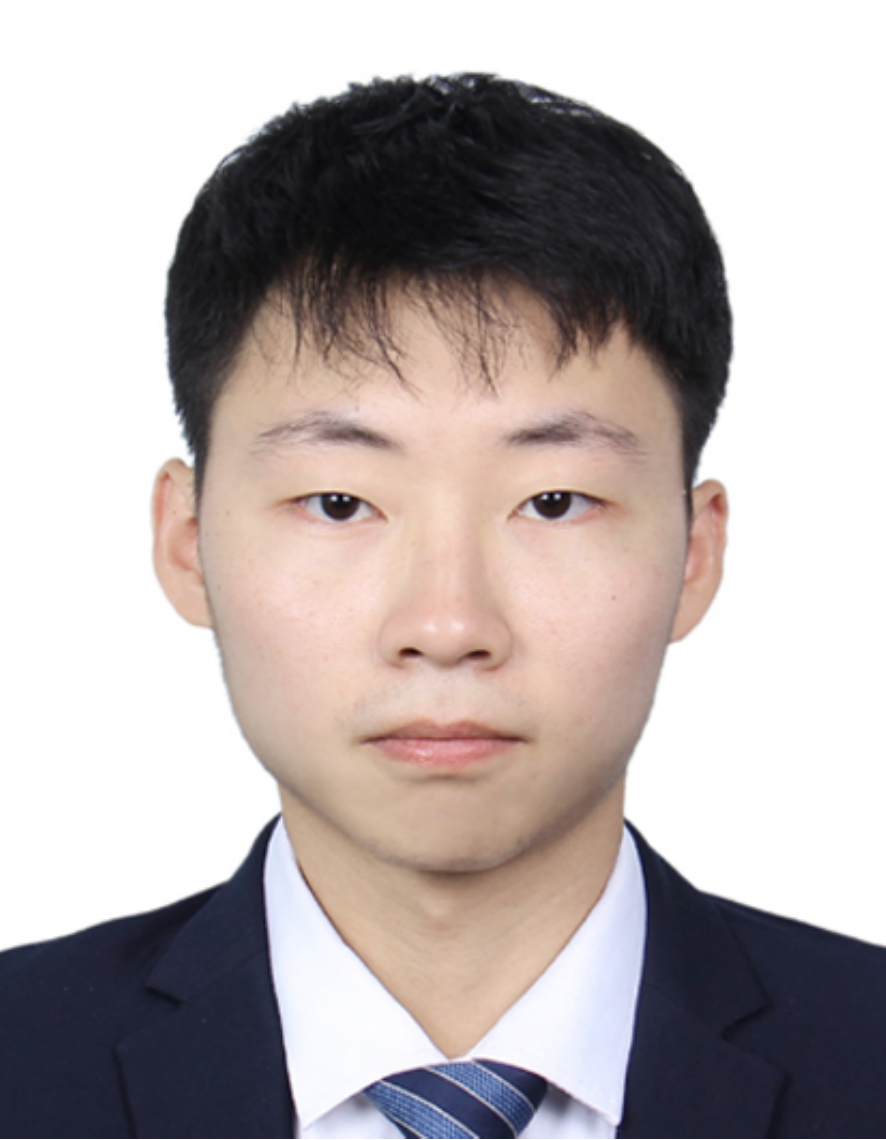}}]{Minghao Lu}
received his Bachelor of Engineering in Automation in 2021 from Harbin Institute of Technology, China. 

He is currently working toward the Ph.D. degree in Mechanical Engineering at Adaptive Robotic Controls Lab (Arc-Lab) from the University of Hong Kong, China. His research interests include motion planning, robotic control, robotic vision, and aerial systems.
\end{IEEEbiography}

\begin{IEEEbiography}[{\includegraphics[width=1in,height=1.25in,clip,keepaspectratio]{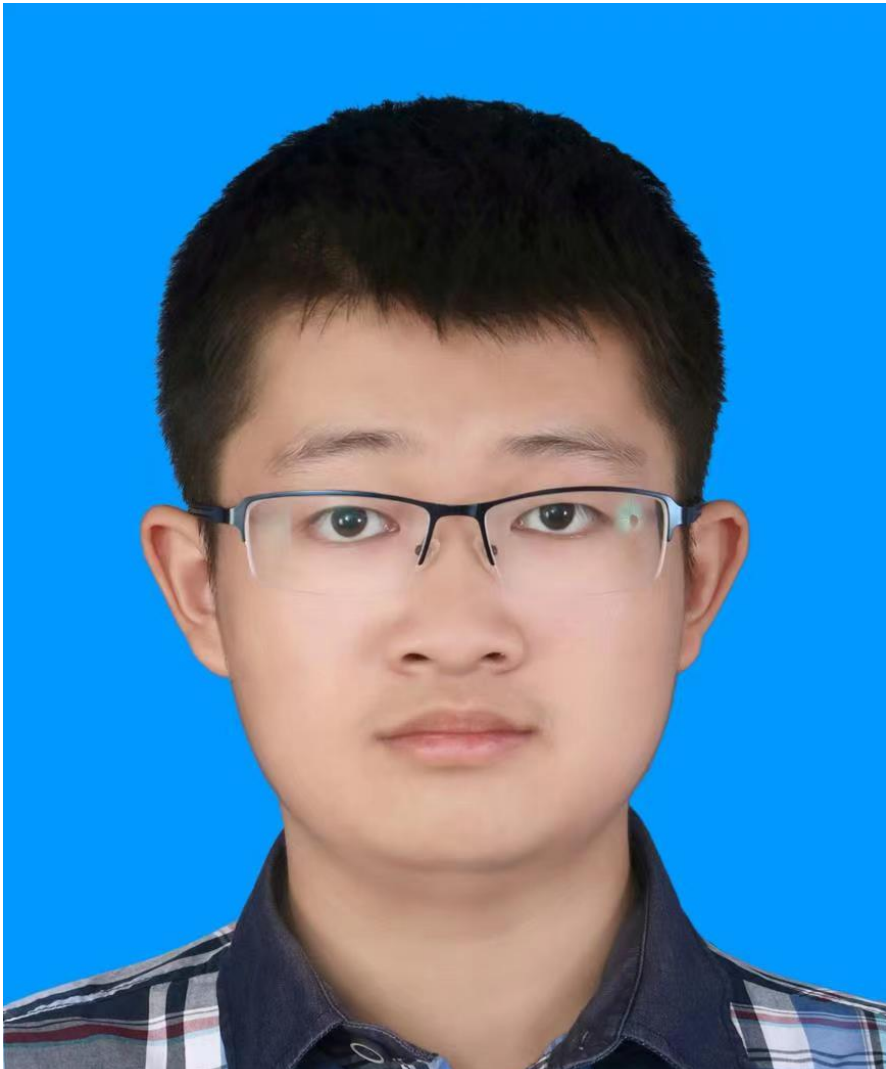}}]{Xiyu Fan}
received his Bachelor of Engineering in Automation in 2023 from Harbin Institute of Technology, China.

He is currently working toward the Ph.D. degree in Mechanical Engineering at Adaptive Robotic Controls Lab (Arc-Lab) from the University of Hong Kong, China. His research interests include reinforcement learning, robotic control, and aerial systems.
\end{IEEEbiography}


\begin{IEEEbiography}[{\includegraphics[width=1in,height=1.25in,clip,keepaspectratio]{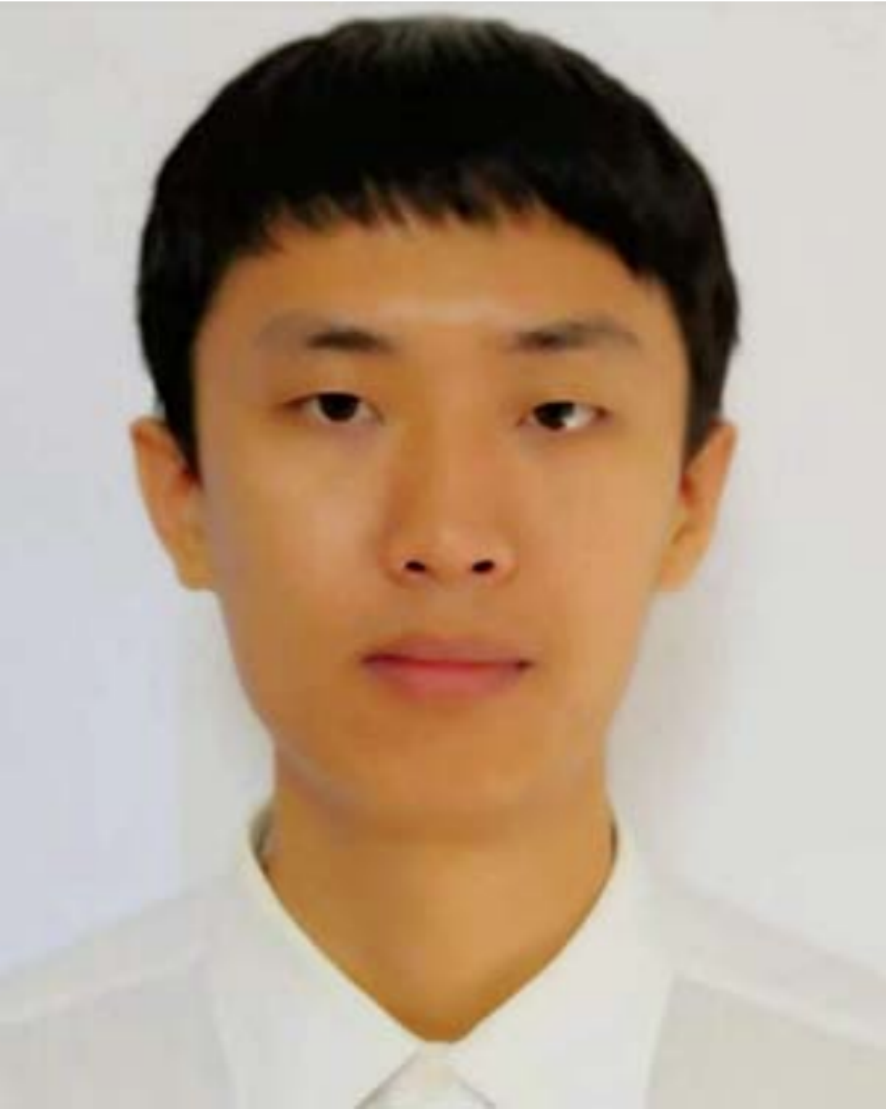}}]{Han Chen}
received his Bachelor of Engineering in 2016 from Beijing Institute of Technology, China, and his Master of Science from Beijing Institute of Technology in 2019. In 2023, he obtained his Doctor of Philosophy from the Department of Aeronautical and Aviation Engineering, The Hong Kong Polytechnic University. 

He is now a senior engineer at Huawei Technologies Co., Ltd, working on mapping engine.
\end{IEEEbiography}

\begin{IEEEbiography}[{\includegraphics[width=1in,height=1.25in,clip,keepaspectratio]{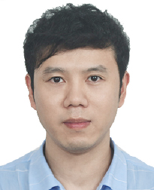}}]{Peng Lu}
obtained his BSc degree in automatic control and MSc degree in nonlinear flight control both from Northwestern Polytechnical University (NPU). He continued his journey on flight control at Delft University of Technology (TU Delft) where he received his PhD degree in 2016. After that, he shifted a bit from flight control and started to explore control for ground/construction robotics at ETH Zurich (ADRL lab) as a Postdoc researcher in 2016. He also had a short but nice journey at University of Zurich \& ETH Zurich (RPG group) where he was working on vision-based control for UAVs as a Postdoc researcher. He was an assistant professor in autonomous UAVs and robotics at Hong Kong Polytechnic University prior to joining the University of Hong Kong in 2020.

Prof. Lu has received several awards such as 3rd place in 2019 IROS autonomous drone racing competition and best graduate student paper finalist in AIAA GNC (top conference in aerospace). He serves as an associate editor for 2020 IROS (top conference in robotics) and session chair/co-chair for conferences like IROS and AIAA GNC for several times. He also gave a number of invited/keynote speeches at multiple conferences, universities and research institutes.
\end{IEEEbiography}




\end{document}